\pdfoutput=1

\documentclass[11pt]{article}

\usepackage[final]{EMNLP2023}

\usepackage{times}
\usepackage{latexsym}
\usepackage{booktabs}
\usepackage[T1]{fontenc}

\usepackage[utf8]{inputenc}

\usepackage{microtype}

\usepackage{inconsolata}
\usepackage{spverbatim}
\usepackage[table]{xcolor}

\title{TASER: Translation Assessment via Systematic Evaluation and Reasoning}

\author{
Monishwaran Maheswaran\textsuperscript{\dag}\textsuperscript{\ddag}\footnotemark[1]\quad
  Marco Carini\textsuperscript{\ddag}\quad
  Christian Federmann \textsuperscript{\ddag}\quad
 Tony Diaz\textsuperscript{\ddag}\\[4pt]
  \textsuperscript{\dag}University of California, Berkeley \quad
  \textsuperscript{\ddag}Apple\\[2pt]
  \texttt{\{monishwaran\}@berkeley.edu},\;
  \texttt{\{m\_carini, chrf, tonydiaz\}@apple.com}
}

\newcommand{\taser}{TASER: Translation Assessment via Systematic Evaluation and Reasoning}

\newcommand{\taserabb}{TASER}

\begin{document}
\maketitle
\footnotetext[1]{Work done while interning at Apple Inc.}
\begin{abstract}
We introduce TASER (Translation Assessment via Systematic Evaluation and Reasoning), a metric that uses Large Reasoning Models (LRMs) for automated translation quality assessment. TASER harnesses the explicit reasoning capabilities of LRMs to conduct systematic, step-by-step evaluation of translation quality. We evaluate TASER on the WMT24 Metrics Shared Task across both reference-based and reference-free scenarios, demonstrating state-of-the-art performance. In system-level evaluation, TASER achieves the highest soft pairwise accuracy in both reference-based and reference-free settings, outperforming all existing metrics. At the segment level, TASER maintains competitive performance with our reference-free variant ranking as the top-performing metric among all reference-free approaches. Our experiments reveal that structured prompting templates yield superior results with LRMs compared to the open-ended approaches that proved optimal for traditional LLMs. We evaluate o3, a large reasoning model from OpenAI, with varying reasoning efforts, providing insights into the relationship between reasoning depth and evaluation quality. The explicit reasoning process in LRMs offers interpretability and visibility, addressing a key limitation of existing automated metrics. Our results demonstrate that Large Reasoning Models show a measurable advancement in translation quality assessment, combining improved accuracy with transparent evaluation across diverse language pairs.
\end{abstract}

\section{Introduction}


Large Language Models (LLMs) have been demonstrated in zero-shot and few-shot translation scenarios, achieving comparable results to dedicated machine translation systems \cite{jiao2023chatgptgoodtranslatoryes, robinson2023chatgptmtcompetitivehigh}. Previous work by \cite{kocmi2023largelanguagemodelsstateoftheart} used Large Language Models (LLMs) through prompting to assess the quality of a machine translation. In their work, GEMBA-DA, they prompt a LLM such as GPT to assess the quality of the translation. Their investigation shows that with straightforward zero-shot prompting, LLMs show accuracy exceeding that of all other non-LLM metrics on the WMT22 \cite{kocmi-etal-2022-findings} evaluation dataset. Their subsequent work, GEMBA-MQM, \cite{kocmi2023gembamqmdetectingtranslationquality} expands on this investigation to detect granular translation quality errors. GEMBA-MQM uses a language agnostic prompting strategy with fixed three-shot prompting to query GPT-4 model to mark error quality spans. Their results indicate GEMBA-MQM achieves state-of-the-art accuracy for system ranking.
In this paper, we introduce {\taserabb}. {\taserabb} builds on these recent findings by investigating Large Reasoning Models. 

Large Reasoning Models \cite{openai2024openaio1card, qwq2024reflectdeeply, deepseekai2025deepseekr1incentivizingreasoningcapability} use long chained reasoning to answer input queries. Reasoning models have shown abilities in problem-solving, coding, as well as scientific reasoning and multi-step logical inference \cite{zhou2025hiddenriskslargereasoning}. Recent findings show that Large Reasoning Models can also be used in translation. \cite{liu2025newtrendsmodernmachine} investigated LRMs at machine translation tasks. In their position paper, they identified three shifts brought about by LRMs: 1) contextual coherence, where LRMs resolve ambiguities and preserve discourse structure through explicit reasoning via context clues; 2) cultural intentionality, where models can adapt translations by inferring speaker intent, audience expectations, and socio-linguistic norms, and finally 3) self-reflection, where LRMs can iteratively refine translations during inference, correcting errors dynamically. These three shifts contribute to more nuanced translations. 

In this paper, we present {\taserabb}. {\taserabb} uses LRMs with zero-shot prompting to arrive at a translation quality estimation. We define and investigate LRMs for the assessment of translation quality in both reference based and reference free scenarios. Starting with the evaluation of the prompts from earlier works that showed state-of-the-art result on non-reasoning LLMs, we iterated on the DA+SQM template used for the human assessment of the translation quality as implemented in the Appraise framework \cite{federmann-2018-appraise} for WMT22 \cite{kocmi-etal-2022-findings} and adapted it towards LRMs. We posit that the strengths of LRMs lead to translation quality estimation that is more aligned with human judgment, as measured in Tables 1 and 2 below. 

The main contributions of this paper are as follows:
\begin{itemize}
    \item We achieve state-of-the-art results using Large Reasoning Models for translation quality assessment on the latest WMT24 \cite{zerva-etal-2024-findings} MQM metrics evaluation dataset.
    \item We evaluate a reasoning model from OpenAI: o3 \cite{o3systemcard} with different reasoning efforts: low and high. Reasoning efforts guide the model on how many reasoning tokens to generate before creating a response to the prompt. Our results show that for translation metric tasks, there isn't any advantage in using high reasoning effort as they both show comparable performance. Performance might however vary, if we had more fine-grained control over the reasoning effort budget.
    \item We conclude that {\taserabb} shows great promise and prompt further investigation into leveraging reasoning models for translation quality assessment.
\end{itemize}

\section{\taserabb \space Metric}
In this method, we prompt reasoning models from OpenAI with the following attributes: source language, target language, source text segment, translation segment, and optionally, the human reference segment, analogous to \cite{kocmi2023largelanguagemodelsstateoftheart}. After iterating and evaluating on different prompts, we observed that simple zero-shot open ended prompting does not result in the best overall assessment. The prompt that we settled on includes the attributes as listed above as well as includes more direction, particularly assessment instructions and details of what to look for during quality assessment. We leave evaluating other reasoning models and additional language pairs for future work.

\section{Experimental Setup}
Our experiments involve measuring the performance of {\taserabb} on the WMT24 Metrics shared task \cite{zerva-etal-2024-findings}, where automated metrics are evaluated against human gold labels. The goal is to predict a quality score for each segment in a given test set which can be a variant of Direct Assessment (DA) or Multidimensional Quality Metrics (MQM). We evaluate {\taserabb} across the evaluation set provided by WMT24. Similar to \cite{kocmi2023gembamqmdetectingtranslationquality}, we compare our method against the best-performing reference-based and reference free metrics of WMT24.

\subsection{Evaluation Datasets}
\textbf{MQM} datasets from the WMT24 \cite{zerva-etal-2024-findings} are across three language pairs: English $\to$ German, English $\to$ Spanish, and Japanese $\to$ Chinese. The dataset contains the source sentences, output of machine translation systems, and reference translations. The quality of each source-translation pair is annotated by at least three independent expert annotators, using DA on a scale 0-100.


\subsection{Evaluation Criteria}
Our evaluation is the same process as the evaluation process followed in \cite{freitag-etal-2024-llms}. 

At the system level, the evaluation is done with soft pairwise accuracy (SPA) \cite{thompson-etal-2024-improving}, which addresses some of the drawbacks of standard pairwise accuracy which does not account for the uncertainty of the system ranking. SPA addresses this problem by using p-values as a proxy for certainty, where p-values are calculated between two systems using both the metric and human scores, then taking 1.0 minus the absolute difference between the two p-values as the metric’s score for that pair, resulting in the same statistical conclusion as the human scores. Moreover, SPA does not reward or penalize metrics with statistical ties rather the accuracy score is proportional to whether or not the metric and human have the same level of certainty in the ranking. 

At the segment level, evaluation follows the same process as \cite{freitag-etal-2024-llms, freitag-etal-2023-results} where pairwise accuracy is computed with tie calibration, that is, metrics are given credit for correctly predicting ties in human scores, while automatically calibrating for each metric's natural scale. The accuracy/correlation scores are then simply averaged for the final score, placing the metric scores on an absolute scale and independent of the performance of other metrics.

\section{Results}
\begin{table}[h!]
\centering
\begin{tabular}{lc}
\hline
\textbf{Metric} & \textbf{SPA}\\
\hline
\rowcolor{lightgray} TASER-o3-low & 0.872 \\
TASER-o3-high & 0.868 \\
\rowcolor{lightgray} TASER-o3-high & 0.867 \\
TASER-o3-low & 0.864 \\
XCOMET & 0.861 \\
MetricX-24-Hybrid & 0.856 \\
MetaMetrics-MT & 0.852 \\
\rowcolor{lightgray} MetricX-24-Hybrid-QE & 0.848 \\
\rowcolor{lightgray} gemba-esa & 0.846 \\
\rowcolor{lightgray} XCOMET-QE & 0.833 \\
COMET-22 & 0.824 \\
BLEURT-20 & 0.821 \\
\rowcolor{lightgray} bright-qe & 0.805 \\
\rowcolor{lightgray} MetaMetrics-MT-QE & 0.802 \\
BLCOM-1 & 0.789 \\
PrismRefMedium & 0.766 \\
PrismRefSmall & 0.760 \\
damonmonli & 0.739 \\
\rowcolor{lightgray} sentinel-cand-mqm & 0.739 \\
YiSi-1 & 0.735 \\
\rowcolor{lightgray} CometKiwi & 0.733 \\
BERTScore & 0.714 \\
chrF & 0.700 \\
MEE4 & 0.696 \\
chrfS & 0.694 \\
spBLEU & 0.671 \\
BLEU & 0.663 \\
sentinel-ref-mqm & 0.570 \\
\rowcolor{lightgray} sentinel-src-mqm & 0.570 \\
\rowcolor{lightgray} XLsimMqm & 0.509 \\
\hline
\end{tabular}
\caption{System level average soft pairwise accuracy (SPA) for all metrics from the WMT24 across the main language pairs: English $\to$ German, English $\to$ Spanish, Japanese $\to$ Chinese. Metrics highlighted gray did not use a reference translation.}
\label{tab:sys_pce}
\end{table}
The results for TASER on the WMT24 test dataset is reported under both reference based and reference free scenarios. The results are compared against the \textbf{MQM} gold labels. TASER is evaluated under two configurations: TASER-o3-low (low reasoning effort setting) and TASER-o3-high (high reasoning effort setting). The low-effort variant corresponds to settings where there are possibly fewer inference steps or less inference time compute as defined by \cite{o3systemcard}, while the high-effort variant leverages more inference time compute. Table 1 reports soft pairwise accuracy (SPA) on the system level scenario averaged across the main language pairs: English $\to$ German, English $\to$ Spanish, Japanese $\to$ Chinese. The results in Table 1 show that TASER achieves the best performance under both reference free and reference based scenarios. The reference-free \textbf{TASER-o3-low} attains state-of-the-art results. The reference based \textbf{TASER-o3-high} outperforms all other metrics including other reference based metrics, only behind reference free \textbf{TASER-o3-low}. Table 2 reports the segment level accuracy with tie calibration. {\taserabb} achieves competitive performance overall with \textbf{TASER-o3-low}, which did not use a reference translation, achieving best overall accuracy among all non reference based metrics. 

\begin{table}[h!]
\centering
\begin{tabular}{lc}
\hline
\textbf{Metric} & \textbf{Accuracy}\\
\hline
MetaMetrics-MT & 0.596 \\
MetricX-24-Hybrid & 0.586 \\
\rowcolor{lightgray} TASER-o3-low & 0.584 \\
\rowcolor{lightgray} TASER-o3-high & 0.584 \\
TASER-o3-high & 0.582 \\
TASER-o3-low & 0.581 \\
\rowcolor{lightgray} MetricX-24-Hybrid-QE & 0.580 \\
\rowcolor{lightgray} gemba-esa & 0.576 \\
XCOMET & 0.576 \\
\rowcolor{lightgray} MetaMetrics-MT-QE & 0.566 \\
\rowcolor{lightgray} sentinel-cand-mqm & 0.560 \\
\rowcolor{lightgray} bright-qe & 0.557 \\
\rowcolor{lightgray} XCOMET-QE & 0.557 \\
COMET-22 & 0.554 \\
BLEURT-20 & 0.550 \\
\rowcolor{lightgray} CometKiwi & 0.547 \\
BLCOM-1 & 0.541 \\
damonmonli & 0.532 \\
PrismRefMedium & 0.526 \\
YiSi-1 & 0.525 \\
PrismRefSmall & 0.524 \\
\rowcolor{lightgray} XLsimMqm & 0.523 \\
BERTScore & 0.522 \\
MEE4 & 0.522 \\
chrfS & 0.520 \\
chrF & 0.516 \\
spBLEU & 0.516 \\
BLEU & 0.515 \\
sentinel-ref-mqm & 0.515 \\
\rowcolor{lightgray} sentinel-src-mqm & 0.515 \\
\hline
\end{tabular}
\caption{Segment level average accuracy with tie calibration for all metrics from the WMT24 across the main language pairs: English $\to$ German, English $\to$ Spanish, Japanese $\to$ Chinese. Metrics highlighted gray did not use a reference translation.}
\label{tab:seg_accuracy}
\end{table}

\section{Conclusion}
In this paper, we introduced \taser, a novel approach that uses Large Reasoning Models (LRMs) for automated translation quality assessment. Our work demonstrates that LRMs can measurably outperform traditional Large Language Models (LLMs) and existing automated metrics in evaluating translation quality. {\taserabb} achieves state-of-the-art performance on the WMT24 Metrics Shared Task when evaluated against the MQM24 dataset. {\taserabb}'s performance demonstrates that the explicit reasoning capabilities of LRMs provide tangible benefits for translation assessment tasks. 

In the near future, we plan to focus on exploring the interpretability advantages offered by the {\taserabb} reasoning process and how they might address the limitations of existing automated metrics. In addition, we plan to investigate {\taserabb} under open-source reasoning models. 

In conclusion, our results suggest that the integration of explicit reasoning processes into evaluation metrics will play a crucial role in advancing the field of machine translation evaluation, ultimately contributing to more reliable and trustworthy automated translation systems across diverse languages and applications.

\section*{Limitations}
{\taserabb} uses off the shelf Large Reasoning Models from OpenAI through prompting. The closed source nature of these models prevent fine-grained control over the reasoning chain and restrict the user from accessing the intermediate reasoning steps, which can limit the interpretability of the model's decision for the quality estimate. Moreover, with off the shelf, closed source models, there is uncertainty on whether models from OpenAI are trained on standard evaluation datasets such as those from WMT24. Therefore, we caution the reader to be mindful of potential data contamination when interpreting the provided results. WMT24 contains a limited set of language pairs which our testing is limited to and results in other language pairs could differ. TASER specific prompts were only used in TASER's evaluation, and were not used in the other LLM-based metrics we compared in Table 1 and 2. Some of the performance we saw could be attributed to the prompt alone. Finally, while LRMs can offer tangible benefits in a variety of tasks, including translation, it does come with increased inference cost when compared to LLMs.




\section*{Acknowledgements}
We acknowledge gracious support from Apple without which this project would not have been completed. The authors are grateful for their peers for their feedback throughout the life cycle of this project. The authors also acknowledge their team's leadership, particularly Adam Archer and Tim Shaw for their invaluable guidance. 

\bibliography{anthology,custom}
\bibliographystyle{acl_natbib}
\appendix
\label{sec:appendix}
\onecolumn

\section{{\taserabb} Prompts}
Below we provide the prompt template used for the experiments described in this paper. There are two prompt templates with minimal variations to account for reference free and reference based scenarios.
\subsection{Reference Free Prompt Template}

\begin{small}
    \begin{spverbatim}
{source_lang} Source: ```{source_seg}```
{target_lang} Machine Translation: ```{target_seg}```
Evaluate the quality of a machine translation for a given segment, using the provided source text, machine-translated text, source language, and target language.

You must analyze the translation without access to any human reference, considering the following:
- Fluency of the translation in the target language.
- Accuracy and completeness of using the information in the source segment.
- Appropriateness of terminology and style for the target language.
- Possible mistranslations, omissions, or additions.
 
Think step by step:
1. First, compare the source and translation for meaning preservation, fidelity, and missing/additional content.
2. Then, analyze fluency, grammar, and naturalness in the target language.
3. Finally, synthesize your findings into a final judgment of quality, including a justification.
 
Continue evaluating as above until all elements have been considered before presenting your final output.
The output should follow this structure:"Score: <your numerical score>"
 
Important: 
- Only use the source and MT segment for evaluation (no references).
- Always provide your reasoning before the final rating and justification.
- Output MUST be valid and must follow the structure.
- Use a continuous scale from 1 (worst) to 100 (best)
\end{spverbatim}
\end{small}

\subsection{Reference Based Prompt Template}
\begin{small}
    \begin{spverbatim}
{source_lang} Source: ```{source_seg}```
{target_lang} Human Reference Translation: ```{reference_seg}``` 
{target_lang} Machine Translation: ```{target_seg}```
Evaluate the quality of a machine translation for a given segment, using the provided source text, human reference translation, machine-translated text, source language, and target language.

You must analyze the machine translation in comparison to the human reference, considering the following:
- Fluency of the translation in the target language.
- Accuracy and completeness of using the information in the source segment and the human reference.
- Appropriateness of terminology and style for the target language.
- Possible mistranslations, omissions, or additions.

Think step by step:
1. First, compare the source and machine translation for meaning preservation, fidelity, and missing/additional content.
2. Then, compare the machine translation with the human reference to analyze fluency, grammar, and naturalness in the target language. 
3. Finally, synthesize your findings into a final judgment of quality, including a justification.

Continue evaluating as above until all elements have been considered before presenting your final output.
The output should follow this structure:"Score: <your numerical score>"

Important: 
- Use the source and MT segment with respect to the human reference for evaluation.
- Always provide your reasoning before the final rating and justification.
- Output MUST be valid and must follow the structure.
- Use a continuous scale from 1 (worst) to 100 (best)
\end{spverbatim}
\end{small}
\end{document}